\documentclass[runningheads]{llncs}
\usepackage[T1]{fontenc}
\usepackage{graphicx}
\usepackage{subcaption}
\usepackage{amsmath}
\usepackage{hyperref}
\usepackage{xcolor,colortbl}

\usepackage[T1]{fontenc}
\usepackage{multirow}
\usepackage{booktabs}
\usepackage{algorithm}
\usepackage{algpseudocode}
\usepackage{amssymb}
\usepackage{booktabs}
\usepackage{tabularx}
\usepackage{longtable}
\usepackage{pifont}

\usepackage{colortbl}
\usepackage[nodisplayskipstretch]{setspace}
\begin{document}
%

\title{DynaMMo: Dynamic Model Merging for Efficient Class Incremental Learning for Medical Images}


%
\titlerunning{DynaMMo: Dynamic Model Merging}
%
\author{Mohammad Areeb Qazi, Ibrahim Almakky, Anees Ur Rehman Hashmi, Santosh Sanjeev, and Mohammad Yaqub}
\authorrunning{Qazi et al.}
%
\institute{Mohamed bin Zayed University of Artificial Intelligence, Abu Dhabi, UAE
\email{\{firstname.lastname\}@mbzuai.ac.ae}}

%
\maketitle              
\begin{abstract}


Continual learning, the ability to acquire knowledge from new data while retaining previously learned information, is a fundamental challenge in machine learning. Various approaches, including memory replay, knowledge distillation, model regularization, and dynamic network expansion, have been proposed to address this issue. Thus far, dynamic network expansion methods have achieved state-of-the-art performance at the cost of incurring significant computational overhead. This is due to the need for additional model buffers, which makes it less feasible in resource-constrained settings, particularly in the medical domain. To overcome this challenge, we propose \textbf{Dyna}mic \textbf{Mo}del \textbf{M}erging, DynaMMo, a method that merges multiple networks at different stages of model training to achieve better computational efficiency. Specifically, we employ lightweight learnable modules for each task and combine them into a unified model to minimize computational overhead. DynaMMo achieves this without compromising performance, offering a cost-effective solution for continual learning in medical applications. We evaluate DynaMMo on three publicly available datasets, demonstrating its effectiveness compared to existing approaches. DynaMMo offers around 10-fold reduction in GFLOPS with a small drop of $2.76$ in average accuracy when compared to state-of-the-art dynamic-based approaches. The code implementation of this work will be available upon the acceptance of this work at: \href{https://github.com/BioMedIA-MBZUAI/DynaMMo}{https://github.com/BioMedIA-MBZUAI/DynaMMo}.

\end{abstract}
\section{Introduction}


The effectiveness of deep neural networks in disease diagnosis has been extensively tested in various settings \cite{zheng2022deep,zhou2021ensembled,zhou2023deep,van2022three,wang2024comprehensive}. However, in practical healthcare scenarios, models are required to continually learn and retain previous knowledge as new data becomes available. This poses a significant challenge, as models often experience forgetting of previously learned information when trained on new data, a phenomenon commonly referred to as \textit{catastrophic forgetting} \cite{french1999catastrophic}. Therefore, a model used for medical diagnosis will only be able to classify the classes on which it was most recently trained. However, a deployable diagnosis system should be able to learn new diseases in a similar manner to a human specialist. The obvious solution to overcoming catastrophic forgetting involves training the model on a combined dataset, including past data. However, this strategy faces many challenges in medical applications due to stringent privacy and security concerns surrounding patient data. Continual Learning (CL) considers such constraints and aims to address catastrophic forgetting, enabling models to adapt to new data while retaining knowledge from previous observations \cite{kirkpatrick2017overcoming,srivastava2021continual}. More specifically, Class Incremental Learning (CIL) is a branch of CL that addresses the process of learning new classes, such as diseases in a medical context. It facilitates the model's ability to continually learn and adapt to evolving data streams while preserving relevant information learned from previous data. In other words, CL aims to aid models in effectively managing changes in the nature of input data while maintaining robust performance over time.


There has been growing interest in addressing the challenges associated with catastrophic forgetting. To achieve this, early CL efforts focused on identifying model parameters crucial to retaining previous knowledge while learning new tasks \cite{kirkpatrick2017overcoming,li2017learning,rebuffi2017icarl}. Techniques such as knowledge distillation have been employed to effectively pass on knowledge from previous models to the most recent one \cite{li2017learning,rebuffi2017icarl}.
However, these methods employ a single model backbone and, therefore, often lack the learning capacity to effectively accommodate new data \cite{grossberg2013adaptive}. Moreover, data replay mechanisms aim to help the model retain what was previously learned by preserving some of the previous data samples for use in later stages of training \cite{boschini2022transfer,buzzega2020dark,rebuffi2017icarl}. While replay-based methods can aid in preserving old knowledge, they encounter challenges in balancing the model's capacity to accommodate new knowledge and the capacity to preserve previously learned knowledge \cite{zhang2023adapter,zhou2023deep}.
On the other hand, Dynamic models have been introduced to dynamically grow a model's ability to accommodate the representations of evolving data streams \cite{douillard2022dytox,wang2022foster}. Such dynamic approaches involve allocating specific parameters for each task. However, the resulting substantial increase in the number of model parameters with every new task leads to storage and computation overheads that are impractical for medical applications. Furthermore, the computational expense associated with storing and utilizing all the backbone structures at inference time presents additional challenges.

Addressing the computational overhead of dynamic CL approaches, light-weight prompt-based methods \cite{douillard2022dytox} and learnable module-based expansions like Adapter-based Continual Learning (ACL) \cite{zhang2023adapter} have gained traction. These methods focus on task-specific features, facilitating accurate classification while maintaining a compact model architecture. Nevertheless, during training and inference, there is a computational overhead associated with passing images through task-specific modules. Additionally, these methods do not facilitate efficient knowledge sharing between task-specific backbones, leading to unnecessary parameter increments. To this end, we introduce Dynamic Model Merging, DynaMMo, a method that achieves a balance between previous and new tasks while maintaining the associated computational costs.
DynaMMo presents a simple yet efficient approach to parameter reduction and utilization during the introduction of new tasks. Furthermore, through DynaMMo, we demonstrate that averaging the parameters of task-specific modules can significantly decrease computational demands without sacrificing performance. In this work, our main contributions can be summarized as follows:

\begin{itemize}
    \item We propose Dynamic Model Merging (DynaMMo), an efficient method for continual class-incremental learning. 
    \item We propose a model merging approach to achieve better computational efficiency during training and inference without compromising on performance.
    \item To the best of our knowledge, DynaMMo is the first method to merge the task-specific components of a dynamic-based continual learning model for medical image analysis. 
\end{itemize}







\section{Related Works}

\subsection{Continual Learning}

Dynamic-based approaches have proven their effectiveness in addressing CIL challenges. They are characterized by their ability to dynamically adjust the model architecture to accommodate new tasks while preserving knowledge from previous tasks. By segregating task-specific parameters from shared parameters, dynamic-based models can learn new tasks without significantly impacting performance on previously learned tasks. Various dynamic strategies have been proposed to expand models for continual learning. One approach involves backbone expansion, where a separate backbone is introduced for each task \cite{yan2021dynamically}. However, this approach is limited by the rapid increase in memory size, making it impractical for scenarios with memory constraints. Another line of research introduces learnable tokens, known as prompts, into the model to facilitate the learning and preservation of new knowledge \cite{douillard2022dytox,wang2022learning}. Nonetheless, this method is primarily applicable to transformer-based models, which typically require substantial amounts of data to achieve generalizability. Recently, \cite{zhang2023adapter} proposed a novel approach for medical tasks, introducing learnable modules called adapters into a pre-trained encoder. These adapters guide the model towards the relevant discriminative features based on the input. However, even though the addition of these adapters does not add a significant amount of additional parameters to the model, they lead to a significant increase in computational complexity. This is due to the number of forward passes that need to be done per input sample. 


\subsection{Model Merging}

Model merging aims to consolidate multiple domain-specific models into a unified model suitable for inference across these domains. 
\cite{wortsman2022model} introduced a method to enhance overall model performance by aggregating the weights of different models that have been selected in a greedy-fashion to improve classification performance. Subsequently, several techniques have been proposed to enhance the performance of model merging \cite{matena2022merging,jin2022dataless,ainsworth2022git,ilharco2022editing,yadav2023resolving,sanjeev2024fissionfusion,almakky2024medmerge}. Model merging has been applied in various scenarios, including federated learning \cite{mcmahan2017communication} and enhancing both out-of-domain \cite{cha2021swad} and in-domain generalization \cite{gupta2020stochastic}. 
In this work, we propose DynaMMo, a method that leverages the classification performance gains of dynamic-based models and then employs model merging to reduce its substantial computational overhead.

\section{Methodology}

\begin{figure}[t]
\centering
\includegraphics[width=\linewidth]{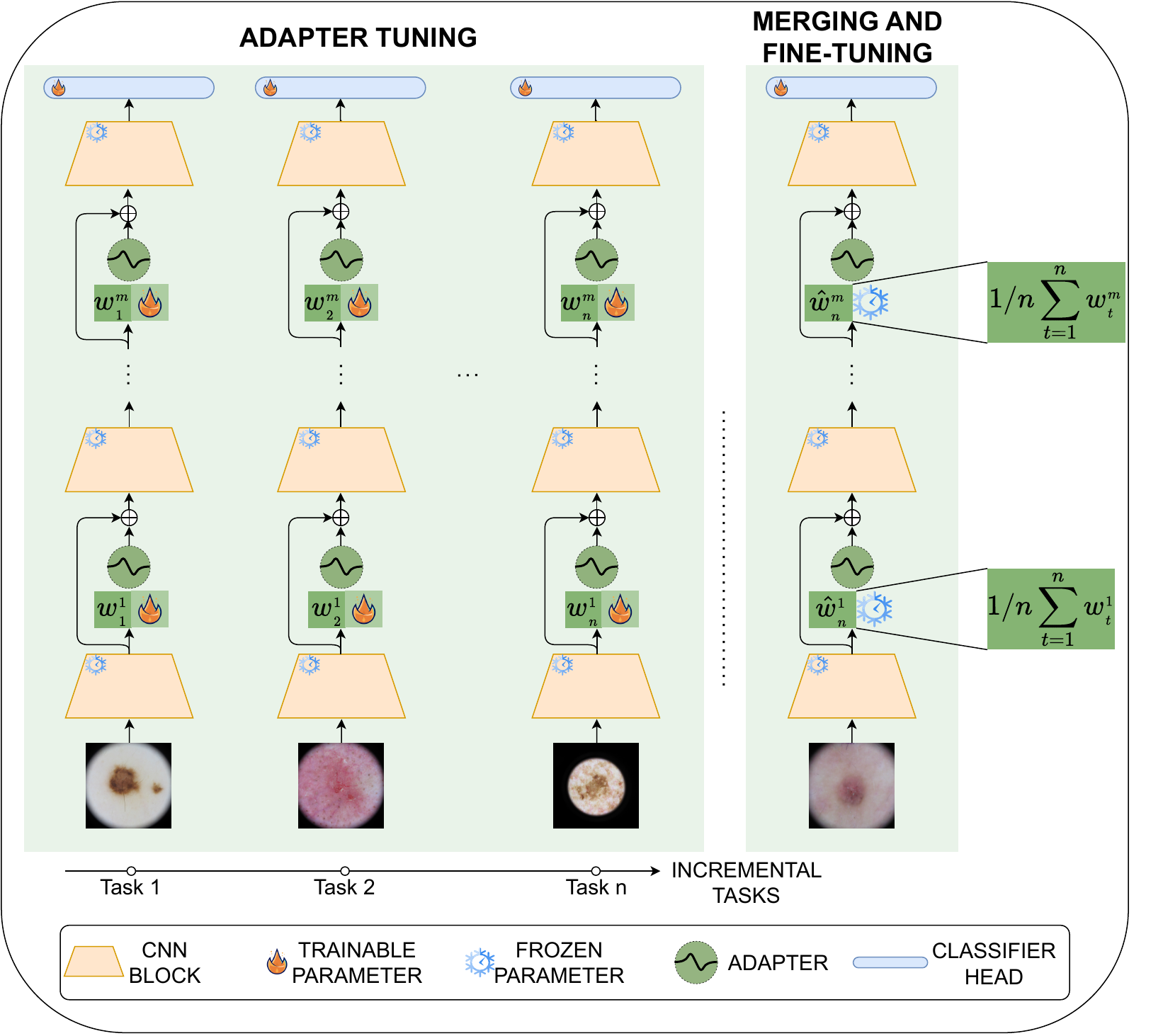}
\caption{Our proposed method, DynaMMo for continual learning, where task-specific adapters are learned to capture task-specific features (Left). Subsequently, these task-specific adapters are merged before fine-tuning the classification head on a balanced set that includes replay data (Right).}
\label{fig:method}
\end{figure}

We propose a Dynamic Model Merging method dubbed DynaMMo, a two-step adapter-based CL method. Firstly with DynaMMo, we ensure the learning of new features specific to the current task using a series of adaptable modules known as adapters. These lightweight learnable modules are incorporated within each block of a Convolutional Neural Network (CNN) model. Optimizing these adapters for each task is referred to as \textit{Adapter Tuning}. To ensure stability and adaptability, a random replay mechanism is employed, whereby a certain number of samples from previous classes are periodically reintroduced during training. Subsequent to learning the new task-specific features, the acquired adapters are merged with the existing ones. Finally, the previous task-specific head is replaced with a unified head using a balanced fine-tuning approach. Figure \ref{fig:method} provides an overview of DynaMMo, which is made up of two stages: Adapter Tuning and Merging and Fine-tuning stages, which we describe in the remainder of this section.

\noindent \textbf{Preliminaries.} The main goal of CIL is to learn a single model that can classify a growing number of classes while retaining its performance on previously learned classes. Let us define a sequence of tasks $\mathcal{D} = \{\mathcal{D}_1, \cdots, \mathcal{D}_n\}$, where every $D_t \in \mathcal{D}$ contains tuples of input samples and their corresponding labels as follows: $D_t = \{(x_i^t, y_i^t)\}_{i=1}^{N_t}$. For any given sample $y_i^t \in C_t$, while $C_t \in \{C_1, \dots, C_n\}$ and all task label sets are mutually exclusive making $C_1 \cap \dots \cap C_n = \phi$. The objective at task $D_t$ is to effectively learn 
from the current data samples without forgetting what was learnt from the previous tasks $\{D_1, \dots, D_{(t-1)}\}$. Furthermore, unlike task incremental settings, in CIL, the sample class is unknown at inference time. 

\subsection{Adapter Tuning}

We start by utilizing the capabilities of a pre-trained CNN backbone, comprised of a set of CNN blocks $\{B_1, \dots, B_m\}$, in extracting effective features for different tasks. We then employ a dynamic-based approach to aid this backbone by training a series of task-specific adapters $\{T_1^1, \dots, T_n^m\}$. 
Each adapter $T_i^j$ has weights $w_i^j$, where $i\leq n$ is the task identifier while $j\leq m$ is the block number. In such manner, $T_i^j$ is dedicated to learning how to adapt the latent representations at every CNN block for every task. These adapters serve to encapsulate the newly acquired task-specific knowledge and integrate it into the model for subsequent tasks. Inspired by \cite{zhang2023adapter}, the adapter architecture starts with a downsampling convolutional layer followed by an upsampling layer. Within each encoder block, the input from the preceding layer is passed through the adapter, generating novel feature representations that are then merged with the original layer's output. This mechanism allows for the progressive acquisition of task-specific features throughout the model. The process of training the task-specific adapters involves training them alongside task-specific heads (TSHs). TSHs are appended to the network and tailored to the number of classes in the respective task along with one other neuron that handles replay samples. By employing Cross-Entropy loss, we simultaneously optimize both the adapters and task-specific heads for each new task. Therefore, the training objective is to minimize the following:


\begin{equation}
\mathcal{L}(\hat{y}, y)=\frac{1}{N_t}\sum_{n=1}^{N_t} -\log \left(\frac{\exp \left(\hat{y}_{n,y_n}\right)}{\sum_{c=1}^{C_{t}} \exp \left(\hat{y}_{n,c}\right)}\right)
\end{equation}

\noindent where $\hat{y}$ is the model prediction based on input $x$, $y$ is the ground truth class for input $x$, $C_t$ is the total number of classes of task $D_t$, and $N_t$ is the total number of samples available for task $D_t$.

The adapter tuning stage offers a flexible and relatively efficient means of extending the functionality of pre-trained CNN models across multiple tasks. By incorporating task-specific knowledge into adaptable modules, we enable the model to continually evolve across various tasks without requiring significant retraining or architectural modifications.

\subsection{Merging and Fine-tuning}


State-of-the-art dynamic models in CL aim to allocate distinct parameters for individual tasks to adapt to changes in the data \cite{douillard2022dytox,yan2021dynamically}. 
More recently, the introduction of task-specific adapters to a shared CNN backbone makes the dynamic approach more efficient \cite{zhang2023adapter}. Nonetheless, such pipelines require multiple forward passes during training and inference to acquire features specific for every task and then passing those features through the task-specific classification heads.  
Consequently, such models suffer from increased computational demands that grows with the increase in the number of tasks. To address this issue, we propose to merge the adapter weights after every adapter tuning step, which allows us to converge to a common minimum. To achieve this, following the adapter fine-tuning process for task $D_t$, we average the task-specific adapters for every CNN block. Therefore, the weights for the task-specific adapter $\hat{w}_t^i$ for task $D_t$ at CNN block $i$ is given by:  

\begin{equation}
\hat{w}_t^i = \frac{1}{t} \sum_{n=1}^{t} w^i_{n}
\end{equation}

The new merged task-specific adapters for task $D_t$ are then frozen and used to fine-tune the expanded classification head for that task. The new classification head incorporates the parameters from the previous task $D_{t-1}$ and is expanded to include the new classes for task $D_t$. We also propose to offset the imbalance issue between the number of samples from the current task and previous tasks by undersampling the number of samples for the current task \cite{Castro_2018_ECCV}. We empirically demonstrate that fine-tuning a single head instead of all the task-specific heads results in improved performance. Our proposed merging process enables the model to classify images encountered thus far with just one forward pass, thereby mitigating computational overhead introduced by adding the task-specific adapters.  
Furthermore, empirical evaluation demonstrates that our proposed merger of parameters reduces the Floating Point Operations Per Second (FLOPS) optimization during training and inference and does not cause significant compromise on performance. By leveraging the collective knowledge encoded in the averaged weights, the model achieves comparable performance to other CL approaches while significantly reducing computational complexity.

\section{Experiments}

\subsection{Datasets}
We use three publicly available datasets to train and evaluate DynaMMo. The three datasets are: CIFAR100 \cite{krizhevsky2009learning}, a widely recognized natural image dataset, PATH16 \cite{zhang2023adapter} a collection of different histopathology datasets, and SKIN8 \cite{tschandl2018ham10000} a skin lesion dataset. Details concerning the number of samples and train/test splits for the three datasets are presented in Table \ref{tab:results}. For CIFAR100, we adopt an incremental CIL approach of 10 classes per task. Whereas for PATH16, we divide the combined dataset into 7 tasks according to the source dataset the images belong to, which are specific in Table \ref{tab:path16}. Finally, for SKIN8, we selected 2 classes per task with a class-stratified train/test split with $\approx16\%$ split ratio. The order of classes for the SKIN8 and CIFAR100 datasets is randomly selected with a set random seed. 
All input images from all datasets were resized to a consistent size of $224\times224$ and normalized to values between $0$ and $1$. During training of all methods, random horizontal flip was applied consistently with a probability of $0.5$.





\begin{table}[t]
\centering
\caption{Details of the datasets used to train, evaluate, and compare the different methods discussed in this work.}
\begin{tabularx}{\textwidth}{@{}>{\centering\arraybackslash}p{2.6cm} *{5}{>{\centering\arraybackslash}X} >{\centering\arraybackslash}X >{\centering\arraybackslash}X >{\centering\arraybackslash}X@{}} 
\toprule
\textbf{Dataset} & \textbf{Classes} & \textbf{Train Set} & \textbf{Test Set} & \textbf{Number of Tasks} \\ \midrule
\textbf{CIFAR100} \cite{krizhevsky2009learning} & 100 & 50000 & 10000 & 10  \\ 
\textbf{PATH16} \cite{zhang2023adapter} & 16 & 12808 & 1607 & 7  \\ 
\textbf{SKIN8} \cite{tschandl2018ham10000} & 8 & 3554 & 704 & 4  \\ 
\bottomrule
\end{tabularx}
\label{tab:results}
\end{table}


\begin{table}[t]
\centering
\caption{Details of PATH16 dataset. This dataset is a combination of 6 different datasets. The split was taken from \cite{zhang2023adapter}.}


\begin{tabularx}{\textwidth}{@{}>{\centering\arraybackslash}p{0.13\textwidth}
                               >{\centering\arraybackslash}p{0.27\textwidth}
                               >{\centering\arraybackslash}p{0.20\textwidth}
                               >{\centering\arraybackslash}p{0.20\textwidth}
                               >{\centering\arraybackslash}p{0.20\textwidth}@{}}

\toprule
\textbf{Task ID} & \textbf{Dataset} & \textbf{Number of Classes} & \textbf{Train Set} & \textbf{Test Set} \\ \midrule
1 & MHIST \cite{wei2021petri} & 2 & 1,000 & 200  \\ 
2 & Breast cancer \cite{cruz2014automatic} & 2 & 1,600 & 200  \\ 
3 & Oral cancer \cite{kebede2021} & 2 & 1641 & 200  \\ 
4 & PatchCamelyon \cite{veeling2018rotation} & 2 & 800 & 200  \\ 
5 & TCGA-STAD \cite{zheng2022deep} & 4 & 3,208 & 407  \\ 
6 & LC25000 \cite{borkowski2019lung} & 2 & 1,600 & 200  \\ 
7 & LC25000 \cite{borkowski2019lung} & 2 & 1,600 & 200  \\ 
\bottomrule
\end{tabularx}
\label{tab:path16}
\end{table}

\subsection{Experimental Setup}
Ensuring fair comparison between all methods, we use ResNet-18 \cite{he2016deep} as the backbone for all methods, including DynaMMo. The ResNet-18 model used was pre-trained on the Imagenet-1K dataset. Opting for the ResNet-18 architecture demonstrates efficacy on small datasets, thereby addressing the challenge of data scarcity prevalent in the medical domain. This is in contrast with Transformer models \cite{dosovitskiy2020image}, which typically require larger datasets for effective generalization.

For DynaMMo, we use Stochastic Gradient Descent (SGD) as the optimizer when training the task-specific adapters, with a batch size of $32$, weight decay of $0.0005$, and momentum of $0.9$. To fine-tune the classification heads, the Adam optimizer was used, with an initial learning rate of $0.001$, which decayed by a multiplicative factor of $0.1$ at the $55^{th}$ and $80^{th}$ epochs. The classifier heads were fine-tuned for $100$ epochs. Following \cite{zhang2023adapter}, the memory limit for storing replay samples was 40 images for SKIN8, 80 images for PATH16, and 2000 images for CIFAR100. We compare DynaMMo with state-of-the-art CIL methods: ICARL \cite{rebuffi2017icarl}, UCIR \cite{hou2018lifelong}, and PODNET \cite{douillard2020podnet}. We also compare DynaMMo against fixed Replay and standard incremental fine-tuning. 



\subsection{Evaluation Metrics}
Following the training process for each task $D_t$, we test the classifier's performance by assessing the accuracy of all classes learned up to that point. The resultant average accuracy is denoted as $\bar{A_t} = \frac{1}{t} \sum_{i=1}^{t} A_{i}$, where $A_i$ is the accuracy at task $D_i$. Additionally, we compare the performance of the last task accuracy denoted as $\bar{A}_n$. Furthermore, to assess the computational efficiency of different methods, we use the number of Floating Point Operations Per Second (FLOPS). 

\section{Results and Discussion}


%

\begin{figure}[t!]
\centering
\includegraphics[width=\linewidth]{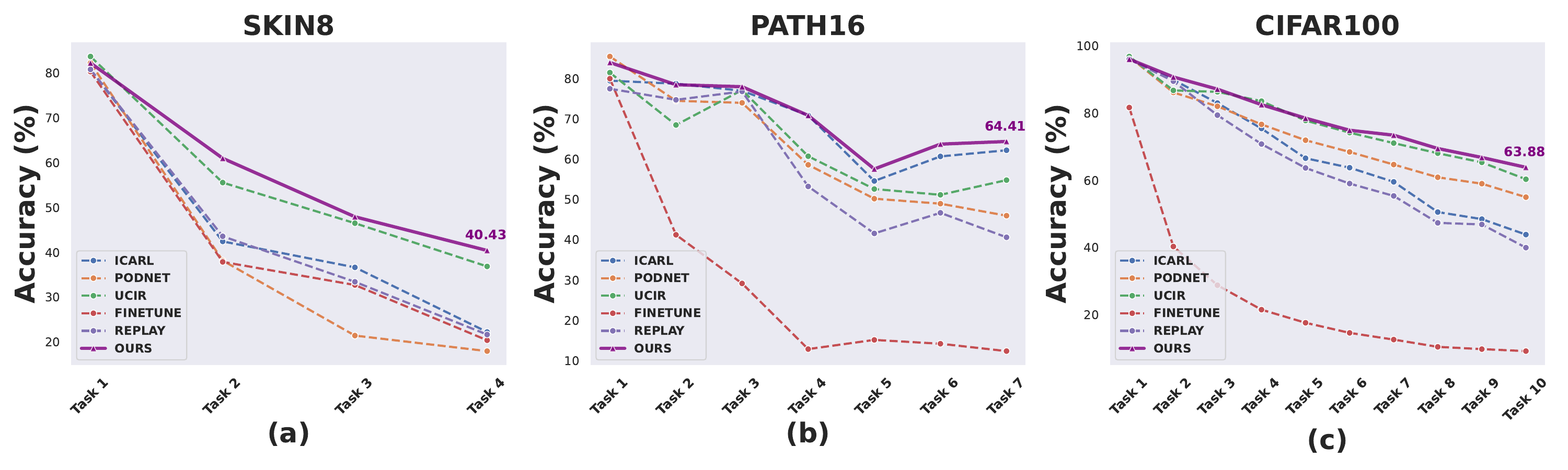}
\caption{Continual Learning Performance comparison between DynaMMo (ours), ICARL \cite{rebuffi2017icarl}, UCIR \cite{hou2018lifelong}, PODNET \cite{douillard2020podnet} along with fixed Replay and standard incremental fine-tuning on (a) SKIN8 \cite{tschandl2018ham10000}, (b) PATH16 \cite{zhang2023adapter}, and (c) CIFAR100 \cite{krizhevsky2009learning} datasets.}
\label{fig:per}
\end{figure}

Figure \ref{fig:per} illustrates the performance of various state-of-the-art CIL methods, including DynaMMo. It is clear that DynaMMo consistently outperforms all other methods. Particularly on the medical datasets, where DynaMMo maintains superior performance throughout the CIL training process. It is worth noting that in Figure \ref{fig:per}, we do not  directly compare ACL \cite{zhang2023adapter} with DynaMMo due to its dynamic network architecture. Dynamic networks, such as ACL, implicitly introduce an additional memory budget in the form of a model buffer for retaining old models. Including this supplementary buffer may lead to biased comparisons with methods that do not store model parameters \cite{zhou2023deep}. Nonetheless, in Table \ref{tab:results_with_acl}, we evaluate the performance drop and computational efficiency (measured in GFLOPS) of ACL against DynaMMo to provide insights into their relative effectiveness. Notably, DynaMMo offers around 10-fold reduction in GFLOPS with a small drop of $2.76$ in average accuracy. DynaMMo achieves this performance with a computational efficiency closer to ICARL \cite{rebuffi2017icarl} and UCIR \cite{hou2018lifelong}, which are regularization based methods.  
Moreover, DynaMMo surpasses the average accuracy of ACL on the SKIN8 dataset, while achieving comparable performance on PATH16. This shows that despite having considerably lower GFLOPS, DynaMMo can outperform dynamic-based models on both natural and medical datasets. Furthrmore, it is important to highlight that unlike ACL, DynaMMo's computational efficiency does not increase with the number of CL tasks. 

\begin{table}[t]
\centering
\caption{
Classification and computational efficiency comparison between DynaMMo and other state-of-the-art approaches. This table shows DynaMMo close to ACL \cite{zhang2023adapter}, in terms of CL classification performance while being closer to ICARL \cite{rebuffi2017icarl} and UCIR \cite{hou2018lifelong} in terms of computational efficiency.  
}
\resizebox{\textwidth}{!}{
\begin{tabular}{ccccc}
\toprule
\textbf{Dataset} & \textbf{Model} & \textbf{GFLOPS$\downarrow$} & \textbf{Avg Acc$\uparrow$} & \textbf{Last Acc$\uparrow$} \\ 
\midrule
\multirow{4}{*}{CIFAR100 \cite{krizhevsky2009learning}} & ACL \cite{zhang2023adapter} & 26.2 & 81.13 & 69.57 \\ 
& \textbf{DynaMMo (Ours)} & 2.62 & 78.37 & 63.88 \\  
& ICARL \cite{rebuffi2017icarl}  & 1.82 & 67.73 & 43.85 \\
& UCIR \cite{hou2018lifelong} & 1.82 & 76.56 & 60.34 \\ 
\midrule
\multirow{4}{*}{PATH16} & ACL \cite{zhang2023adapter} & 18.34 & 76.01 & 71.81 \\ 
 & \textbf{DynaMMo (Ours)} & 2.62 & 71.02 & 64.41 \\ 
& ICARL \cite{rebuffi2017icarl} & 1.82 & 69.76 & 62.23 \\
& UCIR \cite{hou2018lifelong} & 1.82 & 63.79 & 54.82 \\
\midrule
 \multirow{4}{*}{SKIN8 \cite{tschandl2018ham10000}}& ACL \cite{zhang2023adapter} & 10.48 & 57.99 & 43.83 \\
 & \textbf{DynaMMo (Ours)} & 2.62 & 58.01 & 40.43 \\
 & ICARL \cite{rebuffi2017icarl} & 1.82 & 45.54 & 22.27 \\
& UCIR \cite{hou2018lifelong} & 1.82 & 55.65 & 36.88 \\
\bottomrule
\end{tabular}}
\label{tab:results_with_acl}
\end{table}

\begin{table}[t]
\centering
\caption{Ablation study of DynaMMo approach showing the performance difference between different order of steps. The steps are: `AT': Adapter Tuning, `FT': Fine-tuning, `MER': Merging. $^{\dagger}$denotes that the fine-tuning was done using a single classification head rather than using task-specific heads.}
\begin{tabularx}{\textwidth}{@{}>{\centering\arraybackslash}p{2.6cm} *{3}{>{\centering\arraybackslash}X}@{}} 
\toprule
\textbf{Method} & \textbf{CIFAR100} \cite{krizhevsky2009learning} & \textbf{PATH16} \cite{zhang2023adapter} & \textbf{SKIN8} \cite{tschandl2018ham10000} \\ \midrule
\textbf{AT-FT-MER} & 69.42 & 53.53 & 52.65 \\
\textbf{AT-MER-FT} & 77.23 & \textbf{71.02} & 55.58 \\
\textbf{AT-MER-FT$^{\dagger}$} & \textbf{78.37} & 69.38 & \textbf{58.01} \\
\bottomrule
\end{tabularx}
\label{tab:results_ablation}
\end{table}



Table \ref{tab:results_ablation} presents the performance of DynaMMo under different settings. Following the Adapter Tuning step, we explore two configurations: (1) merging the adapters prior to fine-tuning the head and (2) fine-tuning the adapters before merging. Analysis from Table \ref{tab:results_ablation} indicates that merging before fine-tuning consistently yields superior performance across all three datasets. Furthermore, we investigate two approaches for fine-tuning: (1) fine-tuning task-specific heads and (2) conducting single-head fine-tuning. Notably, our findings reveal that a single head suffices to achieve satisfactory performance on the CIFAR-100 and SKIN8 datasets. Conversely, on the PATH16 dataset, task-specific head fine-tuning demonstrates improved efficacy. This is likely because PATH16 is a combination of different datasets and might contain greater domain shifts between the tasks. 

\begin{figure}[t]
    \centering
    \begin{subfigure}{0.33\linewidth}
        \centering
        \includegraphics[width=\linewidth]{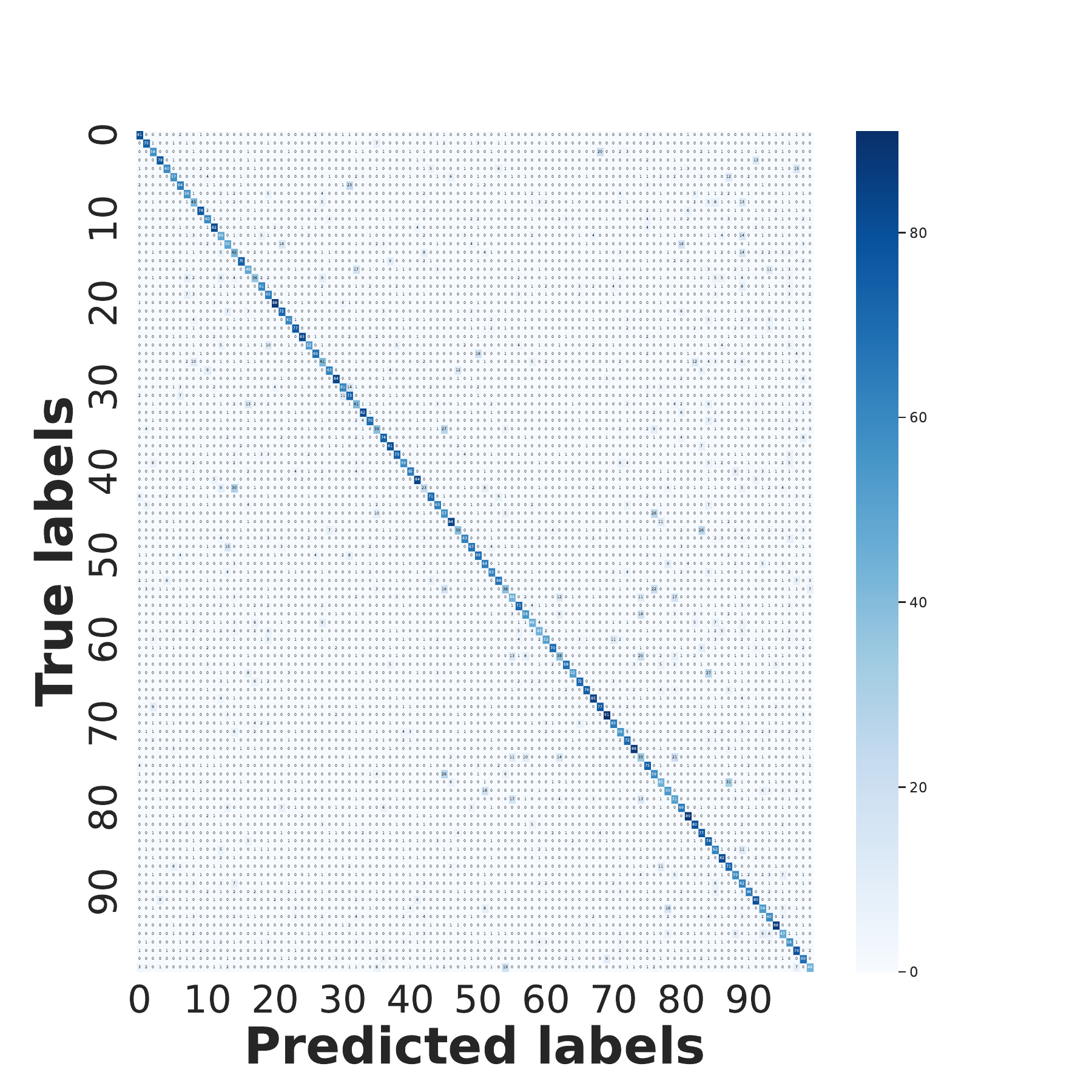}
        \caption{CIFAR100}
        \label{fig:confusion_matrix_cifar100}
    \end{subfigure}%
    \begin{subfigure}{0.33\linewidth}
        \centering
        \includegraphics[width=\linewidth]{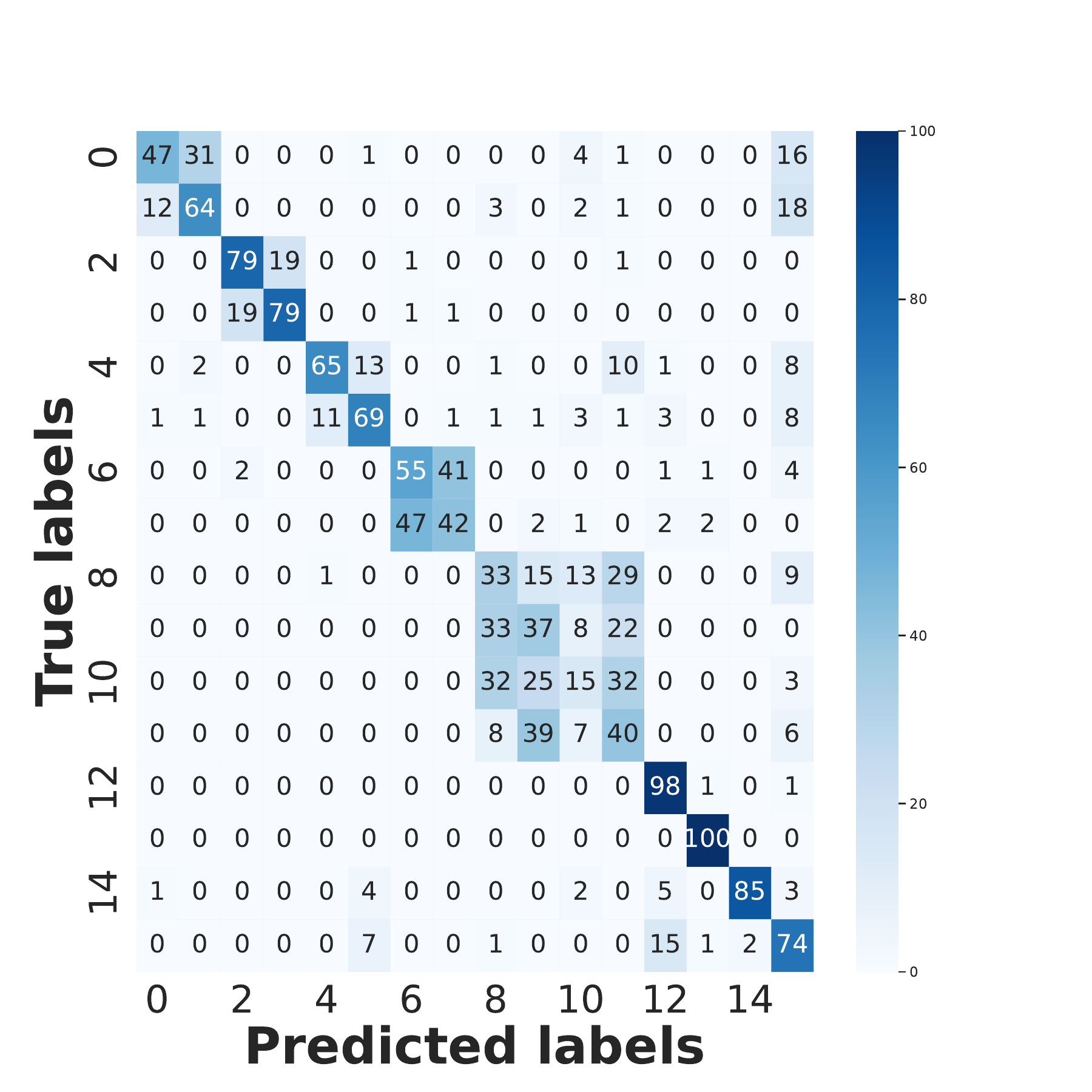}
        \caption{PATH16}
        \label{fig:confusion_matrix_path16}
    \end{subfigure}
    \begin{subfigure}{0.33\linewidth}
        \centering
        \includegraphics[width=\linewidth]{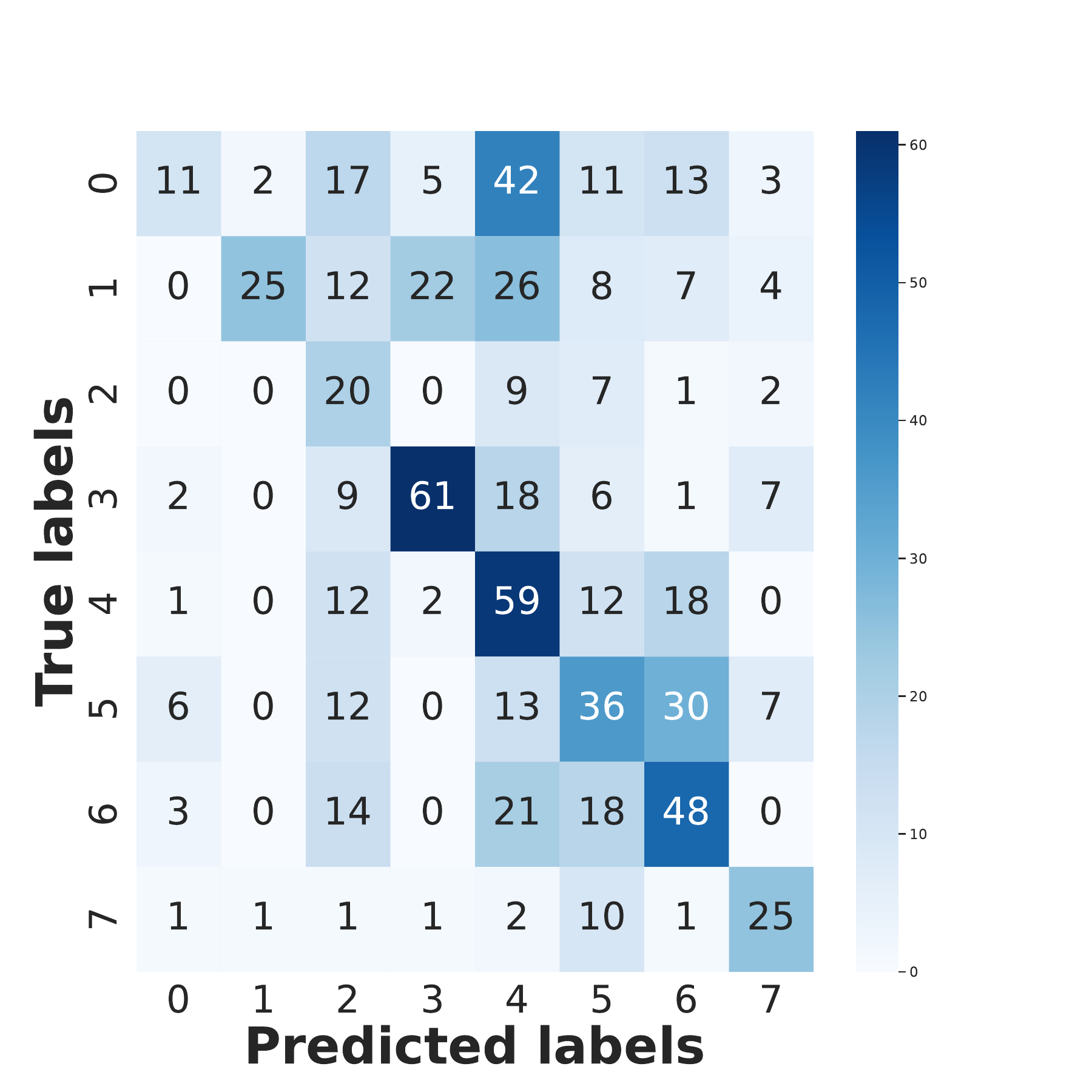}
        \caption{SKIN8}
        \label{fig:confusion_matrix_skin8}
    \end{subfigure}
    \caption{Confusion matrices for DynaMMo on (a) SKIN8 \cite{tschandl2018ham10000}, (b) PATH16 \cite{zhang2023adapter}, and (c) CIFAR100 \cite{krizhevsky2009learning} datasets.}
    \label{fig:cm}
\end{figure}

Figure \ref{fig:cm} illustrates the confusion matrices generated by DynaMMo on all three datasets. The confusion matrices demonstrate the model's ability to distinguish between different classes, including those from the initial task. This observation provides evidence that merging adapter weights does not lead to forgetting classes encountered during previous tasks. The performance depicted in the confusion matrices underscores the robustness and adaptability of DynaMMo across various datasets. By preserving the learnt representations from previous tasks, DynaMMo mitigates the risk of catastrophic forgetting, ensuring effective classification.

\section{Conclusion and Future Work}

In this work, we propose a Dynamic Model Merging method called DynaMMo, which addresses the challenge of catastrophic forgetting and proposes a computationally efficient approach to alleviate it. While dynamic-based CL approaches have achieved superior performance, their application in medical settings has been hampered by computational constraints. To address this challenge, we leverage model merging to harness the strengths of dynamic models while simultaneously reducing the number of parameters and computational efficiency without compromising on performance. Despite having comparable computational efficiency to regularization-based CL methods, DynaMMo outperforms previous state-of-the-art regularization-based techniques in terms of performance. In the future, we plan to investigate other merging methods to achieve better efficiency for continual learning. Moreover, this work focuses on CNN-based backbones, but in the future, we intend to investigate the use of Transformer-based models in medical CL settings. 

\bibliographystyle{splncs04}
\bibliography{main}

\end{document}